\documentclass[preprint,12pt]{elsarticle}

\usepackage{amssymb}
\usepackage{amsmath}
\usepackage{lipsum}
\usepackage{multirow}
\usepackage{algorithm}
\usepackage[noend]{algpseudocode}

\journal{Engineering Applications of Artificial Intelligence}

\usepackage{caption}
\begin{document}

\begin{frontmatter}



\title{Unsupervised Multi-Attention Meta Transformer for Rotating Machinery Fault Diagnosis}


\author[a]{Hanyang Wang}
\author[b,c]{Yuxuan Yang}
\author[d,e,f]{Hongjun Wang\corref{cor1}}
\author[g]{Lihui Wang}
\cortext[cor1]{Corresponding author}

\affiliation[a]{organization={Centre for Effciency and Performance Engineering, University of Huddersfield},
            city={Huddersfield},
            country={HD1 3DH, UK}}

\affiliation[b]{organization={National Key Laboratory of Space Integrated Information System, Institute of Software, Chinese Academy of Sciences},
            city={Beijing},
            country={China}}
\affiliation[c]{organization={Nanjing Forestry University},
            city={Beijing},
            country={China}}

\affiliation[d]{organization={School of Mechanical Electrical Engineering, Beijing Information Science \& Technology University},
            city={Beijing,100192},
            country={China}}
\affiliation[e]{organization={High-end Equipment Intelligent Perception and Control Beijing International Science Cooperation Base},
            city={Beijing},
            country={China}}
\affiliation[f]{organization={Key Laboratory of Modern Measurement \& Control Technology, Ministry of Education, Beijing Information Science \& Technology University},
            city={Beijing},
            country={China}}
            
 \affiliation[g]{organization={Department of Production Engineering, KTH Royal Institute of Technology},
            city={Stockholm},
            country={ Sweden}}           

\begin{abstract}
The intelligent fault diagnosis of rotating mechanical equipment usually requires a large amount of labeled sample data. However, in practical industrial applications, acquiring enough data is both challenging and expensive in terms of time and cost. Moreover, different types of rotating mechanical equipment with different unique mechanical properties, require separate training of diagnostic models for each case. To address the challenges of limited fault samples and the lack of generalizability in prediction models for practical engineering applications, we propose a Multi-Attention Meta Transformer method for few-shot unsupervised rotating machinery fault diagnosis (MMT-FD). This framework extracts potential fault representations from unlabeled data and demonstrates strong generalization capabilities, making it suitable for diagnosing faults across various types of mechanical equipment. The MMT-FD framework integrates a time-frequency domain encoder and a meta-learning generalization model.  The time-frequency domain encoder predicts status representations generated through random augmentations in the time-frequency domain. These enhanced data are then fed into a meta-learning network for classification and generalization training, followed by fine-tuning using a limited amount of labeled data. The model is iteratively optimized using a small number of contrastive learning iterations, resulting in high efficiency. To validate the framework, we conducted experiments on a bearing fault dataset and rotor test bench data. The results demonstrate that the MMT-FD model achieves 99\% fault diagnosis accuracy with only 1\% of labeled sample data, exhibiting robust generalization capabilities.
\end{abstract}


\end{frontmatter}


\section{Introduction}
\label{introduction}

 Rotating machinery, particularly mechanical and electrical equipment, such as gas turbine, aircraft engine, and high-end machining center, represents a critical asset in industrial production. Operating in harsh environments characterized by high temperatures, heavy loads, and continuous operation, this kind of equipment is prone to wear, aging, and failure. Ensuring the safe and stable operation of such equipment is essential for enterprise productivity \cite{gao2015survey},\cite{isermann1997introduction},\cite{wang2022ssd}. Effective fault diagnosis is crucial not only to prevent catastrophic failures, but also to optimize maintenance schedules and reduce operational costs.

Fault diagnosis methods are generally categorized into signal processing-based and data-driven approaches. Signal processing-based methods extract fault-related features from monitoring data using techniques such as Fourier transform (FT), wavelet transform (WT), and empirical mode decomposition (EMD) \cite{moraru2010sensor},\cite{go2016machine}. These methods are particularly effective for analyzing stationary or low-noise signals and can accurately locate faults under ideal conditions. However, their effectiveness is significantly reduced in industrial environments, where vibration signals are often non-stationary and heavily contaminated by noise. Moreover, designing practical signal processing algorithms requires domain expertise and may lack generalizability under different operating conditions.

On the other hand, data-driven methods use machine learning to analyze sensor data for fault detection and classification. Traditional machine learning models, such as support vector machines (SVM), decision trees, and k-nearest neighbors (KNN), rely heavily on accurate feature extraction through time-domain, frequency-domain, or time-frequency domain analysis \cite{blasch2021sensorfusion,mahdavinejad2018iot,wang2024image}. 
Although these methods are robust and interpretable, their performance is heavily dependent on the quality and quantity of labeled data, which is often limited in real-world applications. Additionally, traditional machine learning models are sensitive to variability in operating conditions, leading to challenges in generalization when applied to new or unseen scenarios.

In recent years, deep learning has emerged as a powerful tool for fault diagnosis, demonstrating superior performance in modeling complex patterns and handling large datasets. Popular models such as convolutional neural networks (CNNs) \cite{zhao2008sensordesign,camps2009remotesensing} and autoencoders (AE) \cite{gupta2017predictive} have achieved significant success in various domains. Deep learning's ability to automatically learn hierarchical representations from raw data reduces the dependency on handcrafted feature extraction. However, deep learning still faces several challenges that hinder its broader application in fault diagnosis. First, it requires large volumes of labeled data, which are often scarce in industrial scenarios because of the high cost and time required for data collection and labeling. Second, variability in operating conditions, such as changes in speed and load, leads to domain shifts between the training and testing data distributions, degrading performance \cite{lei2020applications,wang2025advancing}. Finally, the limited availability of fault samples increases the risk of overfitting, particularly for complex models with numerous parameters . These challenges highlight the urgent need for methods that can generalize across variable operating conditions and effectively utilize limited labeled data.

Meta-learning has been introduced as an effective strategy for addressing these challenges \cite{maml,wang2024explicitly,anil,wang2024meta}. By focusing on "learning to learn", meta-learning enables models to adapt rapidly to new tasks with minimal labeled data. This is achieved by leveraging episodic training and situational learning mechanisms, which enhance the model's ability to generalize across diverse tasks \cite{wang2024towards},\cite{protonet}. 
Meta-learning has shown great promise in a few-shot fault diagnosis and cross-domain generalization \cite{reptile}. However, existing meta-learning approaches often struggle with domain shifts, over-fitting due to limited fault samples, and the computational complexity of adapting to diverse industrial scenarios.
Moreover, the integration of domain-specific knowledge, such as time-frequency analysis, into meta-learning frameworks remains an underexplored area.

In this study, we propose MetaTrans, a self-supervised meta-learning framework tailored for fault diagnosis under complex industrial scenarios. MetaTrans incorporates a time-frequency domain encoder and a meta-learning generalization model, effectively addressing the challenges of limited labeled data, domain shifts, and overfitting. The proposed framework leverages time-frequency domain data to extract informative fault features, while self-supervised learning enhances its ability to utilize unlabeled data. By introducing a novel cross-correlation matrix loss and contrastive learning strategies, MetaTrans achieves superior generalization across diverse operating conditions. The contributions are: 
\begin{itemize} \item A self-supervised meta-learning framework is  first to explore for fault diagnosis.It utilizes time-frequency domain data and attention mechanisms, focusing on fault diagnosis with scarce labeled data and varying operating conditions. \item  The MetaTrans with time-frequency multi-attention meta-learning is proposed introducing a cross-correlation matrix and contrastive learning strategies to enhance generalization and reduce reliance on negative samples. \item Comprehensive evaluations on multiple benchmark datasets demonstrate that MetaTrans achieves superior diagnostic accuracy and robustness, outperforming existing state-of-the-art (SOTA) methods. \end{itemize}

In the following, Section 2 shows the related work of the proposed method. Section 3 presents the methodology. Section 4 introduces the rotor system experimental platform. Section 5 presents the experiments fault diagnosis tasks to demonstrate the effectiveness of the proposed method. The final conclusions are made in Section 6.

\section{Related Work}

Fault diagnosis is a critical research area that aims to ensure the safe and efficient operation of industrial equipment by identifying and addressing potential faults. Existing methods for fault diagnosis are typically categorized into two main approaches: signal processing-based methods and data-driven methods.

The signal processing-based method extracts equipment fault-related features from complex monitoring data and then diagnoses potential faults \cite{randall2011vibration}. By enhancing the characteristics of weak fault signals, they are easier to detect and diagnose, improving the sensitivity and accuracy of diagnosis. Common signal processing algorithms include Fourier transform, wavelet transform, short-time Fourier transform, empirical mode decomposition, spectral analysis \cite{mallat2009wavelet},\cite{tang2017bearing}. This type of method can accurately determine the fault and its location, but it is limited by the information obtained and the signal processing method, making it difficult due to the complex working environment and non-stationary vibration signals \cite{jiang2020noise}.

Data-driven fault diagnosis methods have revolutionized fault diagnosis by enabling the automatic extraction of hierarchical features via neural networks from a large amount of mechanical and fault data and then making decisions on new mechanical data . In the field of fault diagnosis, data is obtained based on various sensors, such as acceleration sensors, displacement sensors, temperature sensors, etc. The feature extraction after data collection is crucial, in determining the training and diagnostic performance of subsequent models. Common feature extraction methods include time-domain analysis, frequency-domain analysis, time-frequency domain analysis (STFT Short Time Fourier Transform, Continuous Wavelet Transform CWT, Empirical Mode Decomposition (EMD), etc.) \cite{gonzalez2018time}.

Common time-frequency domain features include energy spectral density, instantaneous frequency, instantaneous bandwidth, etc. Finally, pattern recognition is performed to identify and classify different fault patterns. The related methods can be further divided into traditional methods, e.g., SVM, KNN, Decision Tree, Random Forests \cite{bishop2006pattern}, and deep-learning-based methods that use deep neural networks like CNNs , AE, and RNNs in various fault diagnosis tasks. 

Despite their effectiveness, deep learning methods face several challenges:
(1) they require large amounts of labeled data, which are often unavailable in industrial scenarios;(2) they are susceptible to domain shifts caused by variations in operating conditions, leading to degraded performance; and (3) they risk overfitting due to the limited number of fault samples and the complexity of the models. Recent works have explored domain adaptation and semi-supervised learning to mitigate these issues, yet their adoption in industrial applications remains limited.

Among them, meta-learning has emerged as a promising approach for fault diagnosis. By focusing on ``learning to learn'', meta-learning enables models to adapt to new tasks with minimal labeled data through episodic training and situational learning mechanisms. This approach has shown promise in addressing the limitations of traditional and deep learning methods, particularly in few-shot learning and cross-domain generalization. 

However, meta-learning methods face issues such as overfitting due to limited fault samples, domain shifts, and computational complexity when applied to diverse industrial scenarios. Recent advancements, such as meta-domain adaptation \cite{wang2023awesome}, have attempted to enhance the robustness of meta-learning frameworks by incorporating domain-specific adaptation layers.

Although meta-learning has achieved good performance in few-shot scenarios, challenges remain for real-world industrial applications. Faults in industrial production often occur under diverse operating conditions, and domain shifts significantly impact diagnostic performance. Moreover, some meta-learning methods involve overly complex models that risk overfitting due to the large number of network parameters. Addressing these issues requires innovative approaches that combine domain adaptation, few-shot learning, and meta-learning to develop robust and efficient diagnostic frameworks tailored to industrial environments.
Therefore, the development of robust fault diagnosis methods that can generalize across diverse operating conditions and effectively utilize limited labeled data remains an urgent research priority. In this paper, we aim to leverage meta-learning techniques and propose a multi-attention mechanism combined with time-frequency modulation to achieve robust fault diagnosis.

\section{Methodology}

\begin{algorithm}
\caption{Pseudo-Code of MMT-FD}
\label{alg:mmtfd}
\begin{algorithmic}[1]
\Require Unlabeled dataset $D_u = \{x(t)\}_{t=0}^{T-1}$, distribution of tasks $p(\mathcal{T})$, inner loop learning rate $\alpha$, outer loop learning rate $\beta$, weighting factors $\lambda_1, \lambda_2, \lambda_3$
\Ensure Model parameters $\theta$ initialized for fast adaptation
\For{each unlabeled sample $x(t)$ in $D_u$}
    \State Randomly choose augmentation operations (e.g., window warping, flipping, time-domain noise addition, frequency-domain masking/noise).
    \State Apply augmentation to obtain $x'(t)$.
    \State Convert $x'(t)$ to frequency domain $X'(\omega)$ via FFT.
\EndFor
\For{each augmented sample $x'(t)$}
    \State Extract initial $H_t$ (time) and $H_f$ (frequency).
    \For{$m=1,\ldots,M$ (each attention head)}
        \State $A_m(H_t) = \text{Softmax}\left(\frac{Q_m H_t^T}{\sqrt{d}}\right)H_t$
        \State $A_m(H_f) = \text{Softmax}\left(\frac{Q_m H_f^T}{\sqrt{d}}\right)H_f$
    \EndFor
    \State $H'_t = W_o [A_1(H_t); \ldots; A_M(H_t)]$
    \State $H'_f = W_o [A_1(H_f); \ldots; A_M(H_f)]$
\EndFor
\For{each augmented sample}
    \State $Z_t = \text{Transformer}(H'_t)$
    \State $Z_f = \text{Transformer}(H'_f)$
    \State $Z_t' = G_t(F_t(x'))$, $Z_f' = G_f(F_f(x'))$
    \State Self-Supervised Alignment: Optimize $\mathcal{L}_{align}(Z_t, Z_f)$ to align time and frequency embeddings.
\EndFor
\State Define meta tasks $\{\mathcal{T}_i\}$ from $p(\mathcal{T})$, each with support $(X_i^s, Y_i^s)$ and query $(X_i^q, Y_i^q)$ sets.
\State Introduce classification tasks for time-domain and frequency-domain, optimizing $\mathcal{L}_{cls}^{t}(Z_t)$ and $\mathcal{L}_{cls}^{f}(Z_f)$.
\For{each meta-iteration}
    \State Sample a batch of tasks $\{\mathcal{T}_i\} \sim p(\mathcal{T})$
    \For{each task $\mathcal{T}_i$}
        \State Update
        $\theta_i' = \theta - \alpha \nabla_{\theta}\mathcal{L}_{\mathcal{T}_i}(X_i^s, Y_i^s; \theta)$
    \EndFor
    \State Update
    $\theta \leftarrow \theta - \beta \nabla_{\theta}\sum_i \mathcal{L}_{\mathcal{T}_i}(X_i^q, Y_i^q; \theta_i')$
\EndFor
\State Output $\theta$
\end{algorithmic}
\end{algorithm}

\subsection{Overview of MMT-FD}

In this section, we present the MMT-FD framework, which integrates self-supervised representation learning, meta-learning, and Transformer-based modeling for few-shot fault diagnosis. The core objective of MMT-FD is to learn robust time-frequency representations from unlabeled rotating machinery data and then rapidly adapt to new fault types with only a small amount of labeled examples. The framework consists of four key components:

\begin{enumerate}[(1)]
    \item \textbf{Time-Frequency Data Augmentation:} We design augmentation strategies for both time and frequency domains to alleviate data scarcity and improve generalization. These augmentations introduce diverse variations of the original data, enabling the model to learn robust and transferable features.

    \item \textbf{Multi-Attention Feature Extraction:} By employing multi-head attention mechanisms, the model selectively emphasizes key regions in the time-frequency space. This process highlights discriminative fault-related patterns while reducing the influence of irrelevant noise components.

    \item \textbf{Transformer-based Representation Encoding:} The enhanced time-frequency features are fed into a Transformer encoder. Leveraging self-attention, the Transformer captures global dependencies and contextual information, resulting in comprehensive feature embeddings suitable for few-shot adaptation tasks.

    \item \textbf{Meta-Learning with Bi-level Optimization:} A MAML-based meta-learning approach is integrated. With a meta-learned initialization, the model can rapidly adapt to unseen fault conditions using only a few labeled samples, achieving efficient and effective fine-tuning.
\end{enumerate}

\begin{figure}
    \centering
    \includegraphics[width=1\linewidth]{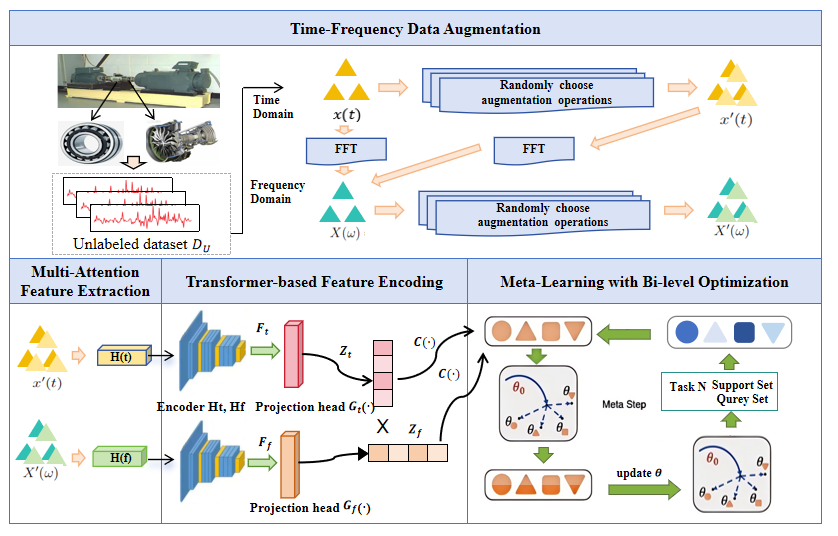}
    \caption{The framework of MMT-FD.}
    \label{fig:MMT-FD}
\end{figure}

By combining self-supervised learning, multi-head attention, Transformer encoding, and meta-learning optimization, MMT-FD extracts robust representations from abundant unlabeled data and quickly adapts to novel scenarios, significantly reducing labeling efforts and improving diagnostic performance in real-world applications, as shown in Fig.\ref{fig:MMT-FD}. Table \ref{tab:symbol} shows the definition of each symbol and Algorithm \ref{alg:mmtfd} shows the pseudo-code of the proposed method.

\subsection{Time-Frequency Data Augmentation}

\begin{figure}
    \centering
    \includegraphics[width=0.7\linewidth]{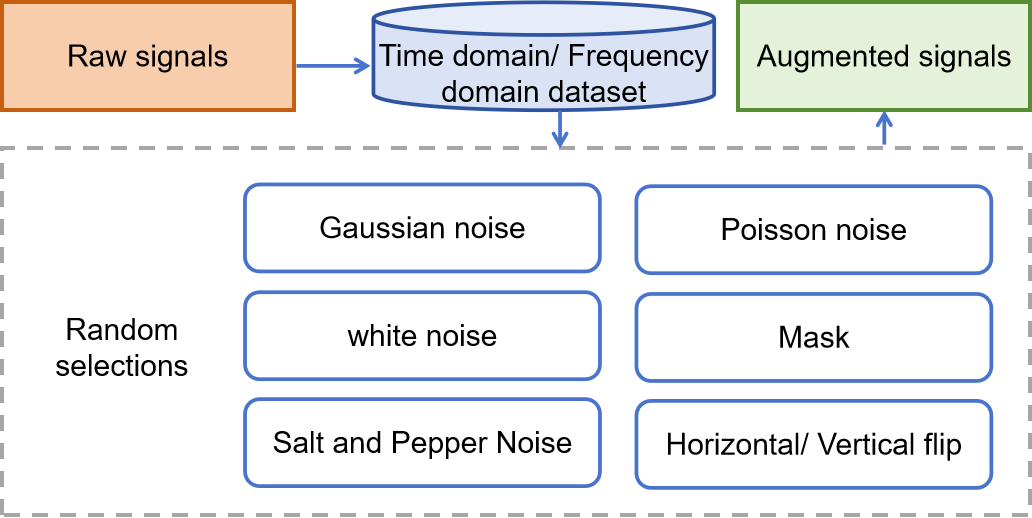}
    \caption{Data augmentation methods.}
    \label{fig:Data Augmentation}
\end{figure}

Data augmentation expands dataset diversity without changing the underlying semantics, which is critical when labeled data is limited. Unlike image-based augmentation (e.g., rotation, cropping, color jittering), time-series augmentation must preserve temporal dependencies and frequency characteristics. To address this, we exploit the properties of rotating machinery signals and develop augmentation strategies in both time and frequency domains, the methods shown in Fig. \ref{fig:Data Augmentation}.

In the time domain, \textit{window warping} can stretch or compress specific time segments without altering the underlying fault characteristics. Let $x(t)$, $t=0,\ldots,T-1$, be the original signal, and consider a segment $(x(t_s), \ldots, x(t_e))$. With a scaling factor $s>0$:
\begin{equation}
x'(t) = 
\begin{cases}
x\left(t_s + \frac{t - t_s}{s}\right), & t \in [t_s, t_s + s(t_e - t_s)]\\[6pt]
x(t), & \text{otherwise}
\end{cases}.
\end{equation}
If $s<1$, the segment is compressed; if $s>1$, it is stretched. Another example is \textit{flipping}, which inverts the amplitude:
\begin{equation}
x'(t) = -x(t).
\end{equation}
Such transformations enrich the training set without altering fault semantics. Additionally, adding Gaussian noise $n(t)\sim \mathcal{N}(0,\sigma^2)$ introduces variability and enhances robustness:
\begin{equation}
x'(t) = x(t) + n(t).
\end{equation}

Transforming $x(t)$ into the frequency domain via the Fast Fourier Transform (FFT) provides further augmentation opportunities. Let $X(\omega)$ be the frequency-domain representation:
\begin{equation}
X(\omega) = \sum_{t=0}^{T-1} x(t)e^{-j 2 \pi \omega t / T}.
\end{equation}
Frequency-domain augmentation may involve masking certain frequency components:
\begin{equation}
X'(\omega) = X(\omega) \cdot M(\omega),
\end{equation}
where $M(\omega)$ is a binary mask, or adding noise in the frequency domain:
\begin{equation}
X'(\omega) = X(\omega) + N(\omega),
\end{equation}
to simulate frequency perturbations and improve generalization.

By randomly selecting and combining multiple augmentation strategies for each training batch, the model encounters a continually shifting distribution of unlabeled but semantically consistent data. This approach ensures the learned representations remain robust and noise-invariant. Such self-supervised augmentation forms a strong foundation for subsequent fine-tuning with limited labeled data.

\subsection{Multi-Attention Mechanism}

After obtaining augmented time-frequency inputs, we need to emphasize salient features and suppress noise. To achieve this, we adopt a multi-attention mechanism that focuses on different time-frequency scales and regions, ensuring that crucial fault patterns are highlighted while reducing the impact of irrelevant information.

Let $H_t \in \mathbb{R}^{L \times d}$ and $H_f \in \mathbb{R}^{L \times d}$ be the time- and frequency-domain representations, where $L$ is the sequence length and $d$ is the embedding dimension. The multi-head attention mechanism consists of $M$ parallel attention heads. Each head uses a query matrix $Q_m$ to attend to different parts of $H \in \{H_t, H_f\}$:
\begin{equation}
A_m(H) = \text{Softmax}\left(\frac{Q_mH^T}{\sqrt{d}}\right)H,\quad m=1,\ldots,M.
\end{equation}

Each attention head captures distinct aspects of the input. The outputs from all heads are concatenated and projected back:
\begin{equation}
H' = W_o [A_1(H); \ldots; A_M(H)],
\end{equation}
where $W_o \in \mathbb{R}^{d \times (M d)}$ is a projection matrix. Applying this multi-attention module to both $H_t$ and $H_f$ leads to richer and more discriminative representations.

\subsection{Transformer-based Feature Encoding and Meta Tasks}

With the enhanced representations $H'_t$ and $H'_f$, we employ a Transformer encoder to capture global dependencies and contextual information. Inspired by TFPred \cite{chen2024tfpred}, we introduce time and frequency encoders ($F_t$, $F_f$) along with projection heads ($G_t$, $G_f$). Through a self-supervised alignment task, the frequency encoder predicts the low-dimensional embeddings of the time-domain signal, ensuring time-frequency consistency and noise-robust representations.

The Transformer encoder processes $H'_t$ and $H'_f$ as:
\begin{equation}
Z_t = \text{Transformer}(H'_t), \quad Z_f = \text{Transformer}(H'_f).
\end{equation}
The self-attention mechanism within the Transformer captures comprehensive features across time and frequency dimensions. The frequency encoder attempts to predict the time-domain representation:
\begin{equation}
Z_f \approx G_f(F_f(X')), \quad Z_t = G_t(F_t(X')),
\end{equation}
where $X'$ denotes the augmented input signal. After pre-training, a classification head $C(\cdot)$ can be attached to $Z_t$ or $Z_f$ and fine-tuned with minimal labeled data for few-shot fault classification.

We introduce multiple meta tasks to enhance generalization:
\begin{enumerate}[(1)]
    \item \textit{Time-Frequency Alignment Task:} Align time and frequency embeddings by having the frequency encoder predict time-domain features.
    \item \textit{Time-Domain Fault Diagnosis Task:} Classify fault types using $Z_t$.
    \item \textit{Frequency-Domain Fault Diagnosis Task:} Classify fault types using $Z_f$.
\end{enumerate}

By jointly optimizing these tasks, the model learns more general, transferable representations suitable for few-shot adaptation to new fault conditions.

\subsection{Meta-Learning Optimization and Final Objective}

To ensure rapid adaptation in few-shot scenarios, we adopt a MAML-based meta-learning framework with a bi-level optimization process: an inner loop for fast adaptation and an outer loop for optimizing meta-parameters.

Let $\theta$ be the model parameters. We sample tasks $\mathcal{T}_i$ from a task distribution $p(\mathcal{T})$, each comprising a support set $(X_i^s, Y_i^s)$ and a query set $(X_i^q, Y_i^q)$. In the inner loop, parameters are updated using only the support set:
\begin{equation}
\theta_i' = \theta - \alpha \nabla_{\theta}\mathcal{L}_{\mathcal{T}_i}(X_i^s, Y_i^s; \theta),
\end{equation}
where $\alpha$ is the inner-loop learning rate. The outer loop updates the meta-parameters based on the query sets:
\begin{equation}
\theta \leftarrow \theta - \beta \nabla_{\theta}\sum_i \mathcal{L}_{\mathcal{T}_i}(X_i^q, Y_i^q; \theta_i'),
\end{equation}
where $\beta$ is the outer-loop learning rate. Repeatedly performing this process across multiple tasks yields an initialization $\theta$ that can quickly adapt to new tasks with minimal labeled data.

The final optimization objective combines the self-supervised alignment loss, classification losses, and meta-learning loss:
\begin{equation}
\mathcal{L}_{final} = \mathcal{L}_{align}(Z_t, Z_f) + \lambda_1 \mathcal{L}_{cls}^{t}(Z_t) + \lambda_2 \mathcal{L}_{cls}^{f}(Z_f) + \lambda_3 \mathcal{L}_{meta},
\end{equation}
where $\lambda_1, \lambda_2, \lambda_3$ are weighting factors. This integrated objective ensures that the model learns robust representations from unlabeled data and adapts efficiently to new fault scenarios with limited labeled samples.

By combining multi-attention, a Transformer-based architecture, and meta-learning optimization, MMT-FD learns resilient, transferable features during self-supervised pre-training and can quickly adapt to new conditions during fine-tuning, ultimately reducing labeling costs and improving fault diagnosis performance in real-world applications.

\begin{table*}[h]
\centering
\caption{Symbol Table}
\label{tab:symbol}
\resizebox{1.0\linewidth}{!}{
\begin{tabular}{c|c}
\hline
Symbol & Meaning \\ \hline
$x(t)$ & Original time-domain signal, where \(t = 0, \ldots, T-1\) represents the time index. \\
$X(\omega)$ & Frequency-domain representation of \(x(t)\). \\
$T$ & Length of the time-domain signal. \\
$X', x', X'(\omega)$ & Augmented signals in the time and frequency domains. \\
$n(t), N(\omega)$ & Noise in the time domain and frequency domain, respectively. \\
$M(\omega)$ & Binary mask for frequency-domain augmentation. \\
$s$ & Scaling factor used in time-domain window warping. \\
$H_t, H_f \in \mathbb{R}^{L\times d}$ & Time- and frequency-domain feature representations. \\
\(L\) & Sequence length of the features.\\
\(d\) & Embedding dimension of the features. \\
$Z_t, Z_f$ & Features obtained after Transformer-based encoding in the time and frequency domains. \\
$F_t, F_f$ & Time and frequency encoders. \\
$G_t, G_f$ & Projection heads used for dimensionality reduction or alignment tasks. \\
$C(\cdot)$ & Classification head for predicting fault types. \\
$M$ & Number of attention heads in the multi-head attention mechanism. \\
$Q_m$ & Query matrix for the \(m\)-th attention head. \\
$\theta$ & Model parameters, including encoders, projection heads, and Transformer layers. \\
$\mathcal{T}_i$ & The \(i\)-th task sampled during meta-learning. \\
$(X_i^s, Y_i^s), (X_i^q, Y_i^q)$ & Support and query sets of the \(i\)-th task. \\
$\alpha, \beta$ & Learning rates for the inner loop and outer loop during meta-learning, respectively. \\
$\lambda_1, \lambda_2, \lambda_3$ & Weights for the alignment, classification, and meta-learning losses. \\
$\mathcal{L}_{align}$ & Alignment loss for ensuring consistency between time- and frequency-domain representations. \\
$\mathcal{L}_{cls}^{t}$ & Classification loss for predicting fault types using time-domain features \(Z_t\). \\
$\mathcal{L}_{cls}^{f}$ & Classification loss for predicting fault types using frequency-domain features \(Z_f\). \\
$\mathcal{L}_{meta}$ & Meta-learning loss for optimizing task generalization across multiple tasks. \\
\hline
\end{tabular}
}
\end{table*}

\section{Rotor System Experimental Platform}
\begin{figure}
   \centering
   \includegraphics[width=\linewidth]{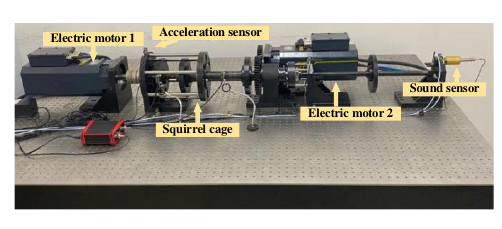}
   \caption{Rotor System Experimental Platform.}
    \label{fig:platform}
\end{figure}

In the study of gas turbine fault diagnosis, we conduct actual tests on a rotor system platform as an essential step. These tests not only provide high-quality experimental data but also validate the effectiveness of theoretical methods and the generalization capability of models under realistic operating conditions. By collecting real vibration signals, we more accurately characterize the state changes of the gas turbine rotor, laying a solid foundation for the development of machine learning and fault diagnosis algorithms. The  platform is shown in Fig. \ref{fig:platform}.

To collect vibration data that reflect different fault states of a gas turbine, we conduct a series of experiments on a gas turbine rotor simulation platform. The core equipment of the platform is an HSP2-30 high-speed servo spindle motor with excellent performance, featuring a rated power of 2.2 kW, a rated speed of 3000 r/min, and a rated torque of 7 N·m. The motor connects to the squirrel cage structure through a coupling in the experimental setup. The squirrel cage includes balance disks on the inner and outer rotors, supported by a gear seat and bearing supports. This layout fully simulates the actual mechanical structure of a gas turbine, ensuring the authenticity and reliability of the experimental data.

In the experiments, we simulate three typical states of the inner rotor: normal condition, imbalance on balance disk 1, and imbalance on balance disk 2. Acceleration sensors installed on the exterior of the squirrel cage structure capture vibration signals under these states. These signals record at a high sampling frequency of 6000 Hz, effectively capturing the high-frequency vibration characteristics of the rotor under different conditions. This provides sufficient information to support subsequent fault diagnosis algorithms.

To expand the dataset and enhance model training, we apply an overlapping sampling method during the data processing phase. Each sample signal length is set to 2048 data points, with a sampling step size of 850 data points, generating numerous samples that cover various states without excessive signal redundancy. This approach ensures the independence of the samples while retaining signal continuity, enhancing the representativeness and robustness of the dataset.

The final dataset divides into training and testing sets in a 7:3 ratio to ensure that the diversity and balance of each state are maintained during model training and validation. Each state generates 214 samples, with 150 samples allocated to the training set and 64 samples to the testing set. Through this meticulous data collection and construction process, we build a dataset that encompasses various fault states, providing a valuable experimental foundation for research into gas turbine fault diagnosis.

\section{Experiments}

\subsection{Datasets}
\label{sec:ex_1}

We select four benchmark datasets for evaluation, including:

\begin{itemize}
    \item The bearing dataset of Case Western Reserve University (CWRU) is among the most commonly utilized datasets in the research on bearing fault diagnosis. This dataset is supplied by the Department of Electrical and Computer Engineering at Case Western Reserve University, which is located in the United States. The dataset contains data related to bearing vibration signals under a range of operating conditions (for instance, load, speed, etc.). It is employed to simulate and study diverse types of bearing faults, among which are inner race faults, outer race faults and rolling element faults. An accelerometer installed on a bearing test bed collects the vibration signal, featuring a high sampling frequency and excellent data quality. The data of each fault type includes multiple fault degrees, providing rich training and test samples, which is suitable for the development and verification of machine learning and deep learning algorithms. 

    \item Paderborn University Bearing Data Center (PUBD) is another important dataset widely used in bearing fault diagnosis research. This dataset contains bearing vibration signal data collected under different operating conditions (such as speed, load, etc.), covering different types of bearing faults and different fault degrees. The vibration signals are collected by various sensors installed on the bearing test equipment, and the data includes normal state, inner ring fault, outer ring fault, rolling element fault and other states. A notable feature of the Paderborn dataset is its detailed experimental setup and data description, which enables researchers to clearly understand the data collection process and operating conditions. 

    \item IMS Bearing Dataset (IMS) is provided by NASA's Intelligent Maintenance System (IMS) program and is an important dataset for bearing fault prediction and diagnosis. The dataset contains bearing vibration signal data collected under long-term operating conditions, recording the whole process of the bearing gradually developing from normal state to fault state. The vibration signals are collected by multiple accelerometers installed on the test equipment. The data collection time is long, providing continuous time series data, which is very suitable for studying the remaining service life prediction and early fault detection of bearings. The IMS dataset contains data on multiple fault types, including inner race faults, outer race faults, and rolling element faults. 

    \item FEMTO Bearing Dataset (FEMTO) is provided by the French FEMTO-ST Institute and is a high-quality dataset for studying the characteristics of bearing faults under different working conditions. The dataset contains bearing vibration signal data collected under different speed and load conditions. The data is collected by multiple sensors installed on the bearing test equipment, covering normal conditions and multiple fault types, such as inner race faults, outer race faults, and rolling element faults. A feature of the FEMTO dataset is its diverse experimental settings and detailed operating condition records, which enable researchers to conduct fault diagnosis studies under different conditions. The dataset is suitable for training and validation of machine learning and deep learning methods, and is helpful for developing and evaluating new fault diagnosis algorithms.

\end{itemize}

\subsection{Compared Baselines}
We compare 12 baseline methods, including a conventional supervised learning method, 3 self-supervised learning methods, 4 meta-learning baseline methods, and 4 state-of-the-art (SOTA) fault diagnosis methods proposed in recent years.
\begin{itemize}
    \item Supervised Learning: This method uses data that has been labeled to train a model, enabling it to forecast issues in a guided way. 
    \item SimCLR \cite{simclr} aims to align the representations of varied versions of the same input. It achieves this through a contrastive loss mechanism and proposes a two-layer non-linear projection head to enhance the quality of the learned representations.
    \item BYOL \cite{byol} involves a dual-stream learning process. One network (online) predicts the representation of the target network for the same input under different transformations, while another is updated based on a moving average of the online network's parameters.
    \item Barlow Twins \cite{barlowtwins} focuses on reducing the redundancy in the representations of modified views. This is achieved by predicting the difference between the representation of a modified view and its noisy version.
    \item MAML is to cultivate model parameters that can effectively adapt to new tasks by utilizing minimal training data and just a few iterations of gradient updates. This approach is designed to be broadly applicable across diverse learning challenges.
    \item Reptile  focuses on optimizing the starting values of neural network parameters. The process involves a cycle of selecting a task, training the model on that task, and then adjusting the initial parameters of the model to be closer to the final trained weights achieved for that specific task.
    \item ProtoNet requires a classifier to adapt to new classes that were not part of the training data, relying on a small number of examples per class. The core of ProtoNet lies in its ability to learn a metric space where the classification is executed by measuring the distances between data points and the prototypical representations of each class.
    \item Relation Network (RelationNet) \cite{relationnet} focuses on developing a sophisticated distance metric through deep learning that enables the comparison of a limited set of images across different learning episodes.
    \item FEDFormer \cite{fedformer} is a Transformer-based deep learning model that effectively captures long-term dependencies in time series data through a self-attention mechanism. In fault diagnosis scenarios, FEDFormer can automatically extract complex features during equipment operation, thereby improving the accuracy of fault detection and prediction.
    \item TimesNet \cite{Timesnet} is a time series analysis model that combines convolutional neural networks and recurrent neural networks. It can capture both local patterns and long-term trends in time series data. In fault diagnosis, TimeNet can identify abnormal patterns in equipment operation and help predict and diagnose potential faults.
    \item TFPred is a time-frequency prediction self-supervised learning framework. It aims to extract latent fault representations from unlabeled fault data. This framework consists of a time encoder and a frequency encoder. The frequency encoder predicts the low-dimensional representation of the time-domain signal generated by the time encoder using randomly augmented data.
    \item MCNN-LSTM \cite{mcnnlstm} is a neural network designed for automatic feature extraction processes raw vibration signals. It employs two convolutional neural networks, each with distinct kernel sizes, to discern varying frequency characteristics from the raw data. Subsequently, long short-term memory (LSTM) is utilized to classify the type of fault based on the features that have been learned.
\end{itemize}

\subsection{Implementation Details}
Our tasks are constructed utilizing image batches, each consisting of $ B = 16 $ images. To enhance the data, we implement the identical data augmentation strategy employed by SimCLR \cite{simclr}, which involves augmenting each image in the batch a total of five times. Essentially, this process involves extracting a random $ 224 \times 224 $ pixel patch from the original signal which is used for fault diagnosis, followed by the application of a series of random augmentations such as horizontal flipping, cropping, and adjustments. For an equitable assessment against various approaches, we employ C4-backbone as the encoders in our study. The sequence of convolutional layers is succeeded by batch normalization, ReLU activation, and max pooling (via strided convolution) in that order. The output of the final layer is then passed to a softmax classifier, which serves as the classification head. These model architectures are pre-trained and remain unchanged throughout the training process. 

Our model undergoes fine-tuning with the Stochastic Gradient Descent (SGD) optimizer. Specifically, the momentum is configured as 0.9, and the weight decay is configured as $10^{-4}$. All experiments are conducted as direct comparisons and executed on NVIDIA RTX 4090 GPUs. For signal processing, we adhere to the conventional setup as outlined in the referenced work by \cite{wang2024towards}. Prior to being input into the model, the data has the mean subtracted from it and is divided by the standard deviation for normalization purposes, and this process is reversed for the output. Our experiments are executed using the PyTorch framework, running on a single NVIDIA 4090 graphics processing unit. We utilize Mean Squared Error (MSE) and Mean Absolute Error (MAE) as the key performance indicators for our evaluations.

\subsection{Evaluation Protocol}
In sequence-based fault diagnosis classification tasks, accuracy (ACC) is adopted as the performance evaluation metric to measure the correctness of the model's predictions for the entire sequence. Specifically, accuracy is defined as the proportion of correctly classified sequences to the total number of sequences. The formula is as follows:
\[
\text{ACC} = \frac{\sum_{i=1}^{N} 1(\hat{y}_i = y_i)}{N}
\]
where \(N\) represents the total number of sequences in the test set, \(\hat{y}_i\) denotes the predicted label for the \(i\)-th sequence, \(y_i\) is the ground truth label for the same sequence, and \(1(\hat{y}_i = y_i)\) is an indicator function that equals 1 if \(\hat{y}_i = y_i\) and 0 otherwise. This metric provides an intuitive measure of the model's overall performance in sequence classification tasks.

\subsection{Performance Comparison}
To comprehensively assess model performance, we construct experiments on the four benchmark datasets described in Subsection \ref{sec:ex_1}. Considering the inclusion of supervised work in the baseline algorithms, we divide the datasets into two groups and use a semi-supervised protocol to ensure that the comparative experiments are conducted objectively. Specifically, we implement the widely accepted protocol, dividing the training dataset into two equally sized subsets by randomly selecting 1\% and 10\% of the data. The models are then further trained for 50 epochs, with distinct learning rates applied to the classifier and the backbone components.

The classifier is fine-tuned with rates of 0.05 and 1.0, while the backbone receives rates of 0.0001 and 0.01, corresponding to the 1\% and 10\% subsets, respectively. It is noteworthy that, based on the approximate viewpoint invariance constructed from the self-supervised learning task, we categorize samples obtained from the same sample augmentation in the self-supervised task as one class, with their pseudo-labels set to the original sample indices.

\begin{figure}
    \centering
    \includegraphics[width=0.7\linewidth]{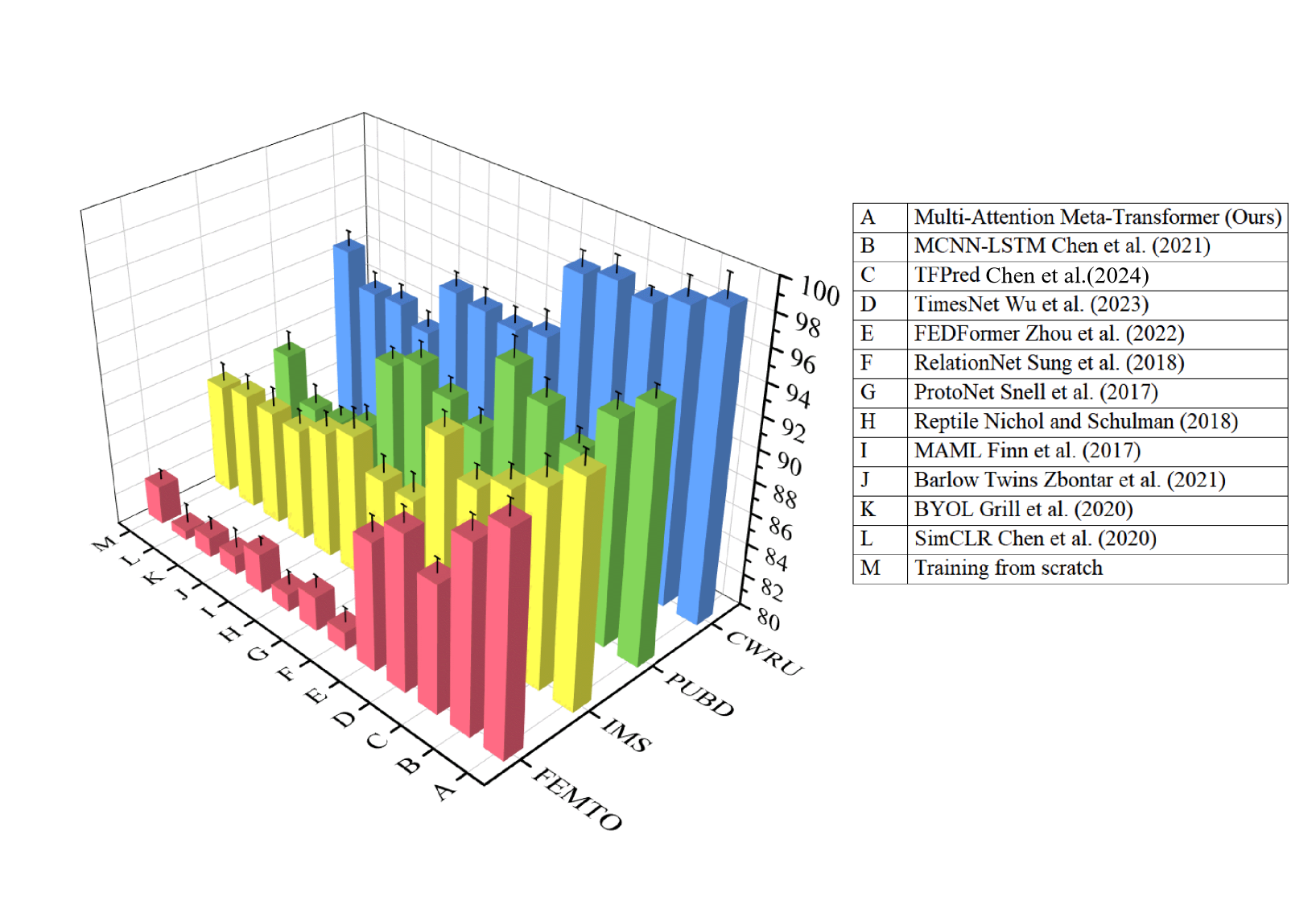}
    \caption{Performance comparison on the four benchmark datasets.}
    \label{fig:P1.1}
\end{figure}

The results are shown in Table \ref{tab:ex_1_1} and Fig. \ref{fig:P1.1}. From the results, we can observe that the proposed method achieves similar or even better results than SOTA. Compared with the self-supervised baseline algorithm, the proposed method achieves an improvement of nearly 4\%, especially on the IMS and FEMTO datasets, achieving a breakthrough of 90\%. This illustrates the superiority of the proposed two-layer optimization framework. Secondly, compared with the meta-learning baseline and supervised learning work, the multi-attention meta-Transformer still achieves similar or even better results than SOTA on all baselines in unsupervised scenarios, achieving an average improvement of 3.8\%.

\begin{table*}[t]
\caption{Performance comparison on the four benchmark datasets. ``Average'' and ``Worst'' denote the average and worst-case accuracy. The best results are highlighted in \textbf{bold}.} 
\setlength\tabcolsep{1pt}
\begin{center}
\resizebox{1.0\linewidth}{!}{
\begin{tabular}{l|cc|cc|cc|cc}
\hline
    \multirow{2}{*}{Method} & \multicolumn{2}{c|}{CWRU}  & \multicolumn{2}{c|}{PUBD}                     & \multicolumn{2}{c|}{IMS}  & \multicolumn{2}{c}{FEMTO}     \\    & Average & Worst 
    & Average & Worst 
    & Average & Worst 
    & Average & Worst  \\ 
\hline
    Training from scratch & 92.51 $\pm$ 0.96 & 90.41 $\pm$ 0.77 & 87.25 $\pm$ 1.21 & 80.36 $\pm$ 1.22 & 87.15 $\pm$ 1.11 & 84.09 $\pm$ 1.23 & 82.51 $\pm$ 0.61 & 77.64 $\pm$ 0.52 \\
    SimCLR \cite{simclr} & 90.51 $\pm$ 1.02 & 88.84 $\pm$ 1.33 & 84.51 $\pm$ 1.22 & 80.52 $\pm$ 1.11 & 87.51 $\pm$ 0.93 & 84.84 $\pm$ 1.01 & 80.60 $\pm$ 1.25 & 75.32 $\pm$ 1.30 \\
    BYOL \cite{byol} & 90.65 $\pm$ 0.91 & 86.74 $\pm$ 1.11 & 84.65 $\pm$ 0.56 & 82.55 $\pm$ 0.66 & 87.38 $\pm$ 1.19 & 85.83 $\pm$ 1.22 & 81.28 $\pm$ 0.77 & 78.38 $\pm$ 0.89 \\
    Barlow Twins \cite{barlowtwins} & 89.65 $\pm$ 0.99 & 88.00 $\pm$ 1.03 & 85.31 $\pm$ 0.99 & 82.01 $\pm$ 1.02 & 87.23 $\pm$ 0.89 & 84.28 $\pm$ 1.17 & 81.22 $\pm$ 1.23 & 79.72 $\pm$ 1.03 \\
    MAML \cite{maml}  & 93.15 $\pm$ 0.88 & 90.51 $\pm$ 1.22 & 90.38 $\pm$ 0.71 & 87.28 $\pm$ 0.91 & 88.10 $\pm$ 0.92 & 85.38 $\pm$ 0.91 & 82.54 $\pm$ 0.54 & 81.84 $\pm$ 0.56 \\
    Reptile \cite{reptile}  & 92.85 $\pm$ 1.00 & 91.84 $\pm$ 1.28 & 91.39 $\pm$ 1.00 & 87.81 $\pm$ 0.92 & 88.98 $\pm$ 1.30 & 84.91 $\pm$ 1.30 & 81.13 $\pm$ 0.89 & 80.93 $\pm$ 0.95 \\
    ProtoNet \cite{protonet}  & 92.56 $\pm$ 1.00 & 90.65 $\pm$ 1.26 & 90.18 $\pm$ 0.92 & 88.09 $\pm$ 1.39 & 87.23 $\pm$ 1.09 & 84.80 $\pm$ 1.00 & 82.19 $\pm$ 0.78 & 79.19 $\pm$ 0.78 \\
    RelationNet \cite{relationnet}  & 93.01 $\pm$ 1.20 & 90.21 $\pm$ 1.13 & 89.18 $\pm$ 0.92 & 88.00 $\pm$ 0.91 & 87.09 $\pm$ 1.21 & 86.89 $\pm$ 1.13 & 81.18 $\pm$ 0.93 & 79.81 $\pm$ 0.93 \\
    FEDFormer \cite{fedformer}  & 97.65 $\pm$ 0.94 & 95.01 $\pm$ 1.02 & 94.18 $\pm$ 1.29 & 91.92 $\pm$ 1.03 & 92.18 $\pm$ 0.92 & 91.10 $\pm$ 0.93 & 88.18 $\pm$ 1.03 & 85.73 $\pm$ 1.18 \\
    TimesNet \cite{Timesnet}  & 98.12 $\pm$ 1.01 & 94.56 $\pm$ 0.99 & 92.81 $\pm$ 1.09 & 87.28 $\pm$ 1.38 & 90.12 $\pm$ 0.87 & 87.89 $\pm$ 1.01 & 89.91 $\pm$ 1.00 & 82.23 $\pm$ 0.99 \\
    TFPred \cite{chen2024tfpred}  & 97.65 $\pm$ 0.52 & 94.98 $\pm$ 0.68 & 91.10 $\pm$ 0.92 & 89.33 $\pm$ 0.94 & 91.23 $\pm$ 0.89 & 90.71 $\pm$ 0.89 & 88.13 $\pm$ 0.90 & 87.19 $\pm$ 0.82 \\
    MCNN-LSTM \cite{mcnnlstm}  & 98.46 $\pm$ 1.19 & 94.65 $\pm$ 1.22 & 94.10 $\pm$ 0.92 & 90.33 $\pm$ 0.96 & 92.45 $\pm$ 1.24 & 91.54 $\pm$ 1.20 & 91.82 $\pm$ 0.93 & 91.00 $\pm$ 0.98 \\
    \textbf{Ours} & \textbf{99.21 $\pm$ 1.44} & \textbf{97.15 $\pm$ 1.06} & \textbf{95.72 $\pm$ 0.78} & \textbf{93.18 $\pm$ 0.82} & \textbf{94.18 $\pm$ 0.95} & \textbf{92.18 $\pm$ 0.98} & \textbf{93.77 $\pm$ 1.01} & \textbf{93.41 $\pm$ 1.03} \\
\hline
\end{tabular}
}
\end{center}
\label{tab:ex_1_1}
\end{table*}

More importantly, compared with the methods used for sequence prediction and fault diagnosis, our proposed method achieves the best results. Meanwhile, faster convergence is achieved based on the meta-learning framework. These results illustrate the effectiveness of the proposed method.

Further, considering that in real-world applications, fault diagnosis models may face various challenges such as noise interference, band acquisition defects, etc., these challenges can lead to the introduction of outliers or the loss of critical data, resulting in decision errors. Therefore, we manually introduce noise into the dataset and compare the effects of different baselines. Specifically, following \cite{wang2024towards}, we add zero-mean Gaussian noise with a variance of 10 to 50\% of the data and apply a mask to 5\% of the region in the complete band. Then, we record the fault diagnosis accuracy of the models.

\begin{figure}
    \centering
    \includegraphics[width=0.7\linewidth]{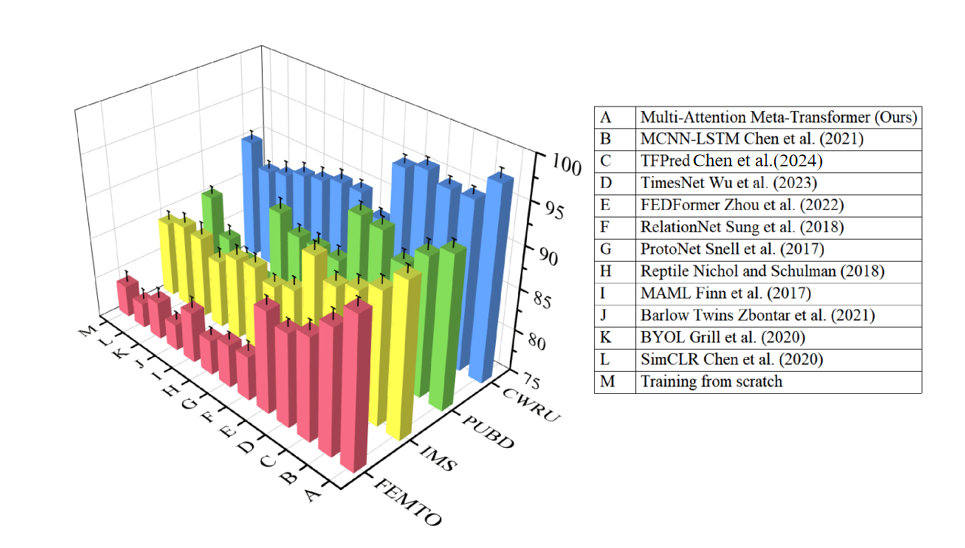}
    \caption{Performance comparison when 50\% of the data are corrupted with Gaussian noise}
    \label{fig:P2}
\end{figure}

Table \ref{tab:ex_1_2} and Fig. \ref{fig:P2} show the results when 50\% of the data are corrupted with Gaussian noise. The results show that our method can still achieve good results even in the presence of noise. At the same time, compared with other baselines, the performance of the proposed method is basically unchanged in the face of noise, with only an average decrease of 1\%. However, the introduction of noise causes the performance of the baseline method to decrease by more than 3\%. These results illustrate the robustness of our method and its advantages in real-world applications.

\begin{table*}
\caption{Performance comparison when 50\% of the data are corrupted with Gaussian noise, i.e., zero mean with the variance of 10. ``Average'' and ``Worst'' denote the average and worst-case accuracy. The best results are highlighted in \textbf{bold}.} 
\setlength\tabcolsep{1pt}
\begin{center}
\resizebox{1.0\linewidth}{!}{
\begin{tabular}{l|cc|cc|cc|cc}
\hline
    \multirow{2}{*}{Method} & \multicolumn{2}{c|}{CWRU}  & \multicolumn{2}{c|}{PUBD}                     & \multicolumn{2}{c|}{IMS}  & \multicolumn{2}{c}{FEMTO}     \\    & Average & Worst 
    & Average & Worst 
    & Average & Worst 
    & Average & Worst  \\ 
\hline
    Training from scratch & 89.89 $\pm$ 1.02 & 86.12 $\pm$ 1.08 & 85.13 $\pm$ 0.89 & 77.21 $\pm$ 0.89 & 84.22 $\pm$ 1.03 & 82.81 $\pm$ 0.93 & 78.87 $\pm$ 1.32 & 75.39 $\pm$ 1.28 \\
    SimCLR \cite{simclr} & 87.18 $\pm$ 0.92 & 83.13 $\pm$ 1.09 & 80.89 $\pm$ 1.23 & 79.23 $\pm$ 1.18 & 85.09 $\pm$ 1.13 & 81.41 $\pm$ 1.15 & 78.00 $\pm$ 1.22 & 75.30 $\pm$ 0.92 \\
    BYOL \cite{byol} & 87.81 $\pm$ 1.29 & 85.32 $\pm$ 1.33 & 79.18 $\pm$ 1.23 & 77.19 $\pm$ 1.22 & 84.83 $\pm$ 1.48 & 82.10 $\pm$ 1.39 & 79.55 $\pm$ 1.21 & 75.88 $\pm$ 1.20 \\
    Barlow Twins \cite{barlowtwins} & 88.89 $\pm$ 1.12 & 85.12 $\pm$ 1.09 & 78.10 $\pm$ 0.92 & 77.10 $\pm$ 0.92 & 83.18 $\pm$ 1.13 & 81.13 $\pm$ 1.09 & 78.18 $\pm$ 0.93 & 74.39 $\pm$ 1.00 \\
    MAML \cite{maml}  & 89.39 $\pm$ 1.12 & 87.80 $\pm$ 1.28 & 87.89 $\pm$ 1.23 & 86.30 $\pm$ 1.19 & 84.83 $\pm$ 1.13 & 82.93 $\pm$ 1.02 & 81.12 $\pm$ 1.30 & 80.91 $\pm$ 1.29 \\
    Reptile \cite{reptile}  & 90.23 $\pm$ 1.09 & 88.90 $\pm$ 1.21 & 86.18 $\pm$ 0.93 & 86.00 $\pm$ 0.97 & 85.19 $\pm$ 1.03 & 81.90 $\pm$ 1.00 & 79.18 $\pm$ 0.93 & 78.19 $\pm$ 1.05 \\
    ProtoNet \cite{protonet}  & 90.28 $\pm$ 0.91 & 86.49 $\pm$ 0.87 & 86.06 $\pm$ 1.20 & 84.55 $\pm$  $\pm$ 1.16 & 84.19 $\pm$ 0.87 & 81.13 $\pm$ 0.89 & 80.17 $\pm$ 0.86 & 78.15 $\pm$ 0.95 \\
    RelationNet \cite{relationnet}  & 88.13 $\pm$ 0.81 & 85.87 $\pm$ 0.91 & 85.89 $\pm$ 1.13 & 83.18 $\pm$ 1.12 & 85.18 $\pm$ 0.92 & 82.38 $\pm$ 0.79 & 80.31 $\pm$ 1.15 & 78.96 $\pm$ 1.29 \\
    FEDFormer \cite{fedformer}  & 95.28 $\pm$ 0.93 & 91.38 $\pm$ 0.97 & 92.38 $\pm$ 0.95 & 90.11 $\pm$ 1.00 & 90.52 $\pm$ 1.02 & 89.20 $\pm$ 1.23 & 87.23 $\pm$ 1.00 & 84.55 $\pm$ 1.02 \\
    TimesNet \cite{Timesnet}  & 96.12 $\pm$ 0.89 & 93.18 $\pm$ 0.92 & 92.00 $\pm$ 0.89 & 86.80 $\pm$ 0.92 & 88.39 $\pm$ 1.01 & 86.39 $\pm$ 0.99 & 86.18 $\pm$ 0.93 & 81.99 $\pm$ 0.92 \\
    TFPred \cite{chen2024tfpred}  & 95.13 $\pm$ 0.87 & 93.32 $\pm$ 0.90 & 89.38 $\pm$ 1.09 & 85.89 $\pm$ 1.13 & 89.38 $\pm$ 1.02 & 86.81 $\pm$ 1.08 & 87.32 $\pm$ 1.15 & 85.00 $\pm$ 1.16 \\
    MCNN-LSTM \cite{mcnnlstm}  & 95.19 $\pm$ 0.82 & 92.45 $\pm$ 0.90 & 91.65 $\pm$ 1.21 & 89.31 $\pm$ 0.87 & 90.77 $\pm$ 0.95 & 90.01 $\pm$ 1.00 & 89.91 $\pm$ 1.23 & 85.81 $\pm$ 1.15 \\
    \textbf{Ours} & \textbf{98.00 $\pm$ 1.05} & \textbf{96.01 $\pm$ 1.02} & \textbf{93.18 $\pm$ 0.92} & \textbf{90.89 $\pm$ 1.00} & \textbf{93.28 $\pm$ 0.89} & \textbf{91.60 $\pm$ 0.92} & \textbf{92.77 $\pm$ 0.95} & \textbf{90.38 $\pm$ 0.95} \\
\hline
\end{tabular}
}
\end{center}
\label{tab:ex_1_2}
\end{table*}

\subsection{Ablation Study}

\begin{figure}
    \centering
    \includegraphics[width=0.7\linewidth]{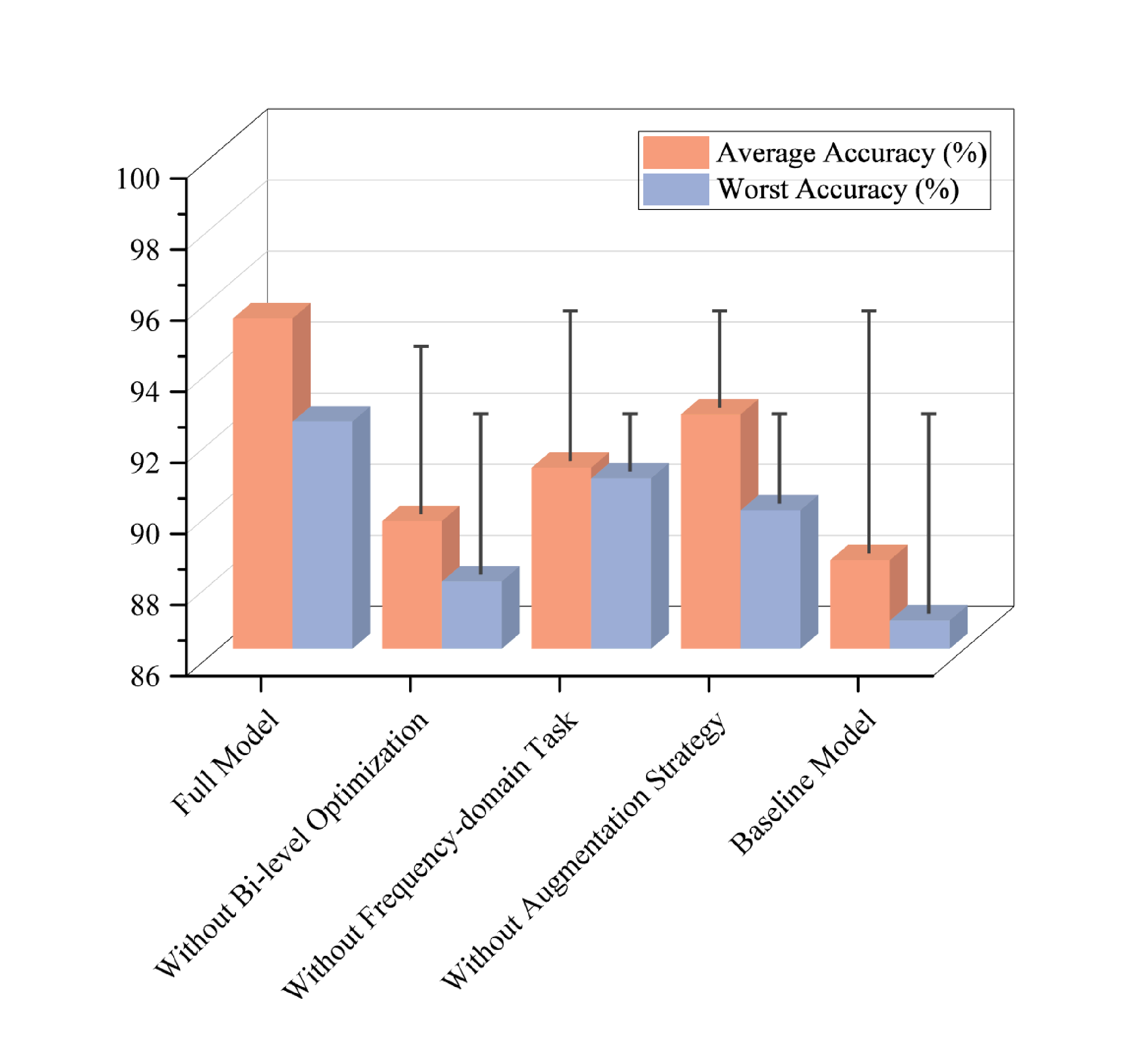}
    \caption{Ablation study results of key modules on IMS dataset.}
    \label{fig:P3}
\end{figure}

To evaluate the effectiveness of our method, we designed and conducted ablation experiments to progressively verify the contribution of key modules to the model's performance. These key modules include augmentation strategies, frequency-domain auxiliary tasks, and a dual-level loop mechanism. By individually removing or replacing these modules, we observed the changes in the model’s performance, thereby confirming the importance of each module. In each experiment, we recorded performance metrics such as model accuracy. The experimental results are shown in Table \ref{tab:abla} and Fig.\ref{fig:P3}. From the results, we can observe that the dual-level optimization mechanism and the frequency-domain auxiliary tasks are critical to the model’s performance, with the model performance decreasing by nearly 5\% when these modules are removed. This indicates that these modules play a crucial role in the overall architecture. Additionally, experiments on the augmentation mechanism show that different strategies impact model performance by approximately 2.7\%. The ablation results further validate the significant advantage of our proposed method over the baseline model, demonstrating its effectiveness and necessity in addressing the target task.

\begin{table*}
\centering
\caption{Ablation study results of key modules on IMS dataset.}
\label{tab:abla}
\resizebox{1.0\linewidth}{!}{
\begin{tabular}{c|c|c|c|c}
\hline
\textbf{Experiment Setup}     & \textbf{Average Accuracy (\%)} & \textbf{Performance Change (\%)} & \textbf{Worst Accuracy (\%)} & \textbf{Performance Change (\%)} \\ \hline
Full Model  & 95.3  & - & 92.4 & - \\ \hline
Without Bi-level Optimization & 89.6 & -4.7 & 87.9 & -4.5 \\ \hline
Without Frequency-domain Task   & 91.1   & -4.2 & 90.8 & -1.6   \\ \hline
Without Augmentation Strategy   & 92.6                   & -2.7        & 89.9 & -2.5 \\ \hline
Baseline Model                & 88.5                   & -6.8            & 86.8 & -5.6 \\ \hline
\end{tabular}}
\end{table*}

\subsection{Visualization}

\subsubsection{Signal Waveform Visualization}
\begin{figure}
    \centering
    \includegraphics[width=0.7\linewidth]{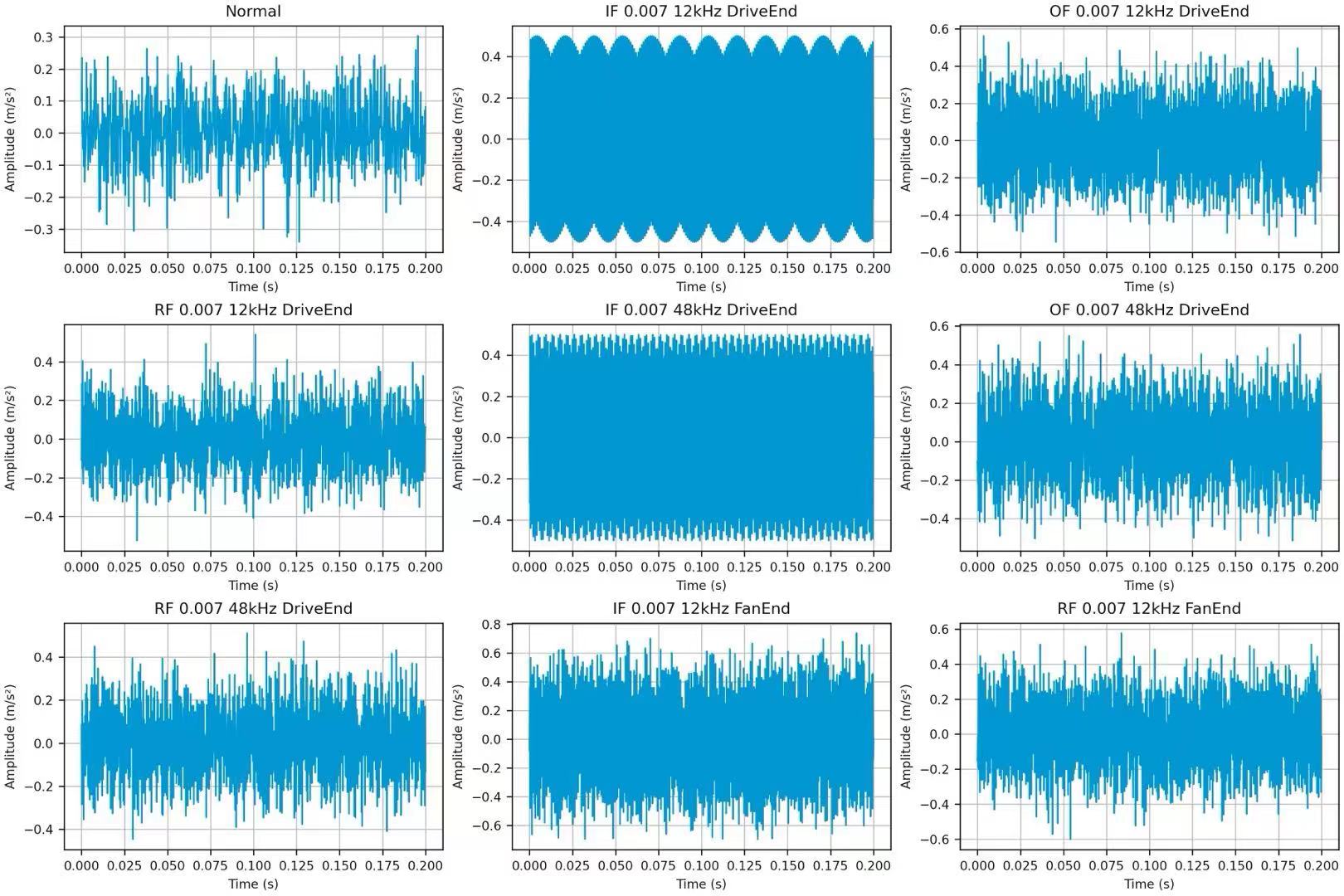}
    \caption{Visualization of reconstructed signal waveforms under various fault conditions. Our method demonstrates higher fidelity in preserving key features compared to baseline methods.}
    \label{fig:waveform}
\end{figure}

Reconstructing and visualizing signal waveforms is crucial to evaluate whether the model can preserve essential features of the original signals. Signal fidelity directly correlates with the model’s ability to diagnose faults accurately under various conditions. In this experiment, we test reconstructed waveforms under conditions such as "IF 0.006 12kHz DriveEnd" and "RF 0.006 48kHz FanEnd." As shown in Fig.\ref{fig:waveform}, MMT-FD exhibits higher fidelity in reconstructed signals compared to baseline methods. This demonstrates the robustness of our time-frequency domain encoder in capturing key signal features.

\subsubsection{Feature Visualization}
\begin{figure}
    \centering
    \includegraphics[width=0.7\linewidth]{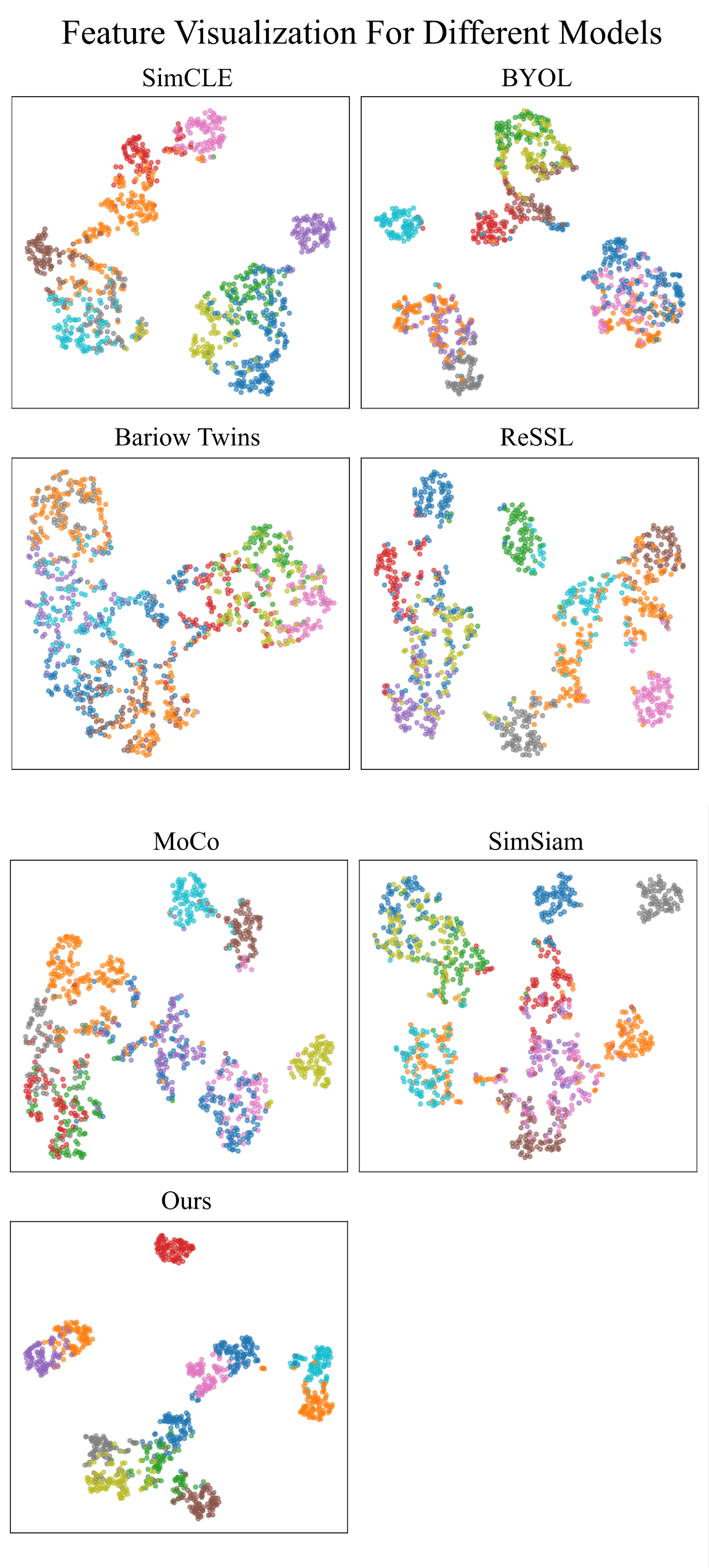}
    \caption{t-SNE visualization of feature embeddings learned by various self-supervised learning models. Our method demonstrates superior clustering and separability compared to the baselines.}
    \label{fig:tsne}
\end{figure}

Visualizing feature embeddings is essential for understanding how well a model captures the underlying structure of the data. Feature embeddings that form distinct, well-separated clusters indicate effective representation learning, which directly impacts generalization performance. We use t-SNE to project the high-dimensional embeddings into a two-dimensional space, allowing for a clear comparison of clustering quality across models. As shown in Fig.~\ref{fig:tsne}, our method demonstrates superior clustering and separability compared to the baseline methods. The compact and distinct clusters produced by MMT-FD confirm its effectiveness in learning robust fault representations.

\subsubsection{Confusion Matrices}
\begin{figure}
    \centering
    \includegraphics[width=0.7\linewidth]{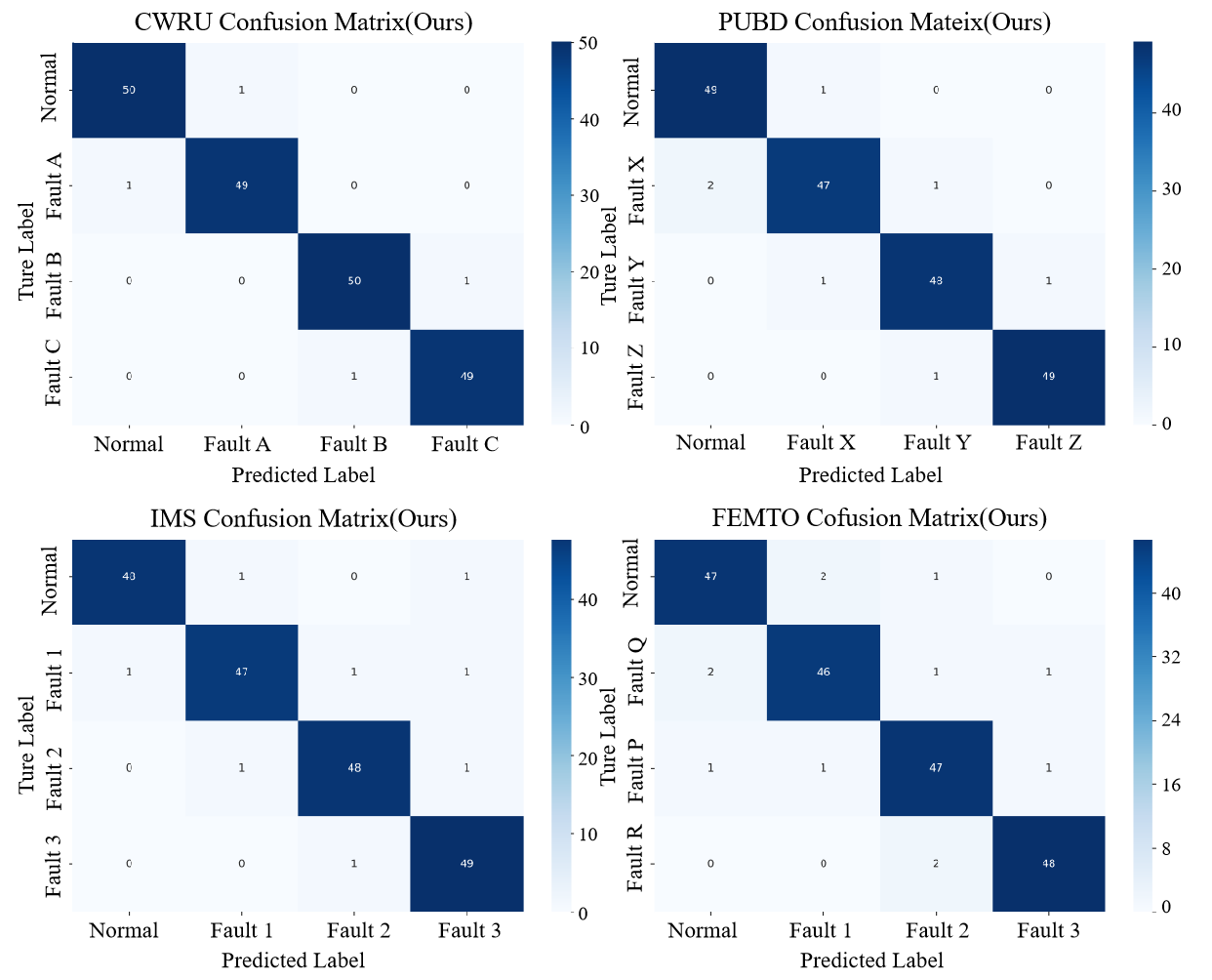}
    \caption{Confusion matrices for fault diagnosis on multiple datasets. Our method consistently achieves higher true positive rates and lower false positive rates, demonstrating superior diagnostic accuracy and generalization.}
    \label{fig:confusion}
\end{figure}

Confusion matrices provide a detailed evaluation of classification performance, showing true positive, false positive, and misclassification rates for each fault type. This visualization is necessary to understand how well a model generalizes across diverse datasets and fault types. We evaluate MMT-FD and baseline methods on CWRU, PUBD, IMS, and FEMTO datasets. As shown in Fig.~\ref{fig:confusion}, our method consistently achieves higher true positive rates and lower false positive rates, confirming its reliability and accuracy in fault diagnosis.

\subsubsection{Loss vs Iteration}
\begin{figure}
    \centering
    \includegraphics[width=0.7\linewidth]{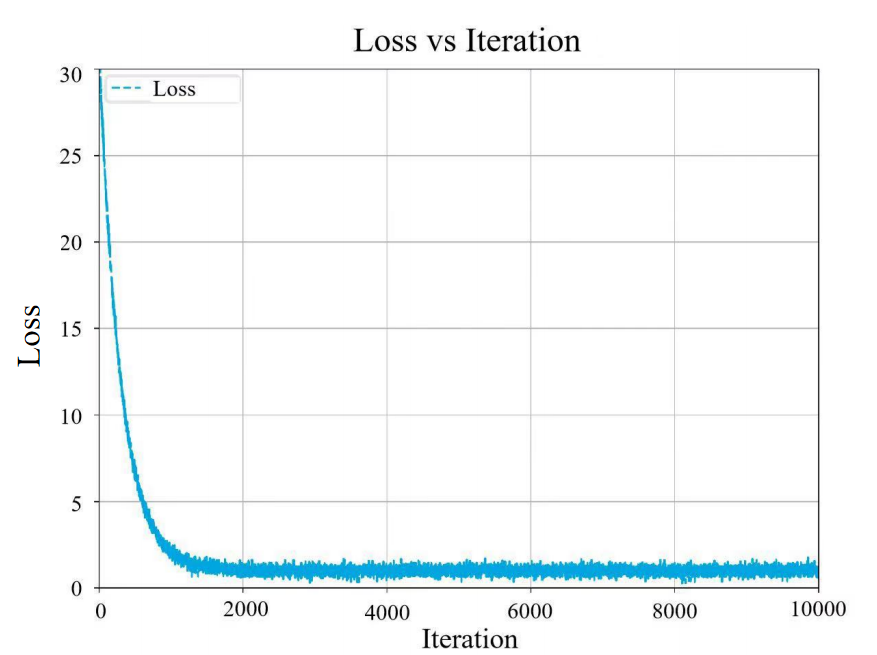}
    \caption{Training loss versus iterations. Our method converges faster and achieves a lower final loss, demonstrating superior optimization efficiency.}
    \label{fig:loss}
\end{figure}

Analyzing the loss curve during training provides insights into the convergence speed and stability of different methods. A model that converges faster and achieves a lower final loss is more efficient and stable in its optimization process. This experiment is necessary to compare how well each method learns representations under limited labeled data conditions. As shown in Fig.~\ref{fig:loss}, MMT-FD achieves a faster convergence rate and a lower final loss compared to the baselines, validating its superior optimization performance.

\subsubsection{Training Accuracy}
\begin{figure}
    \centering
    \includegraphics[width=0.7\linewidth]{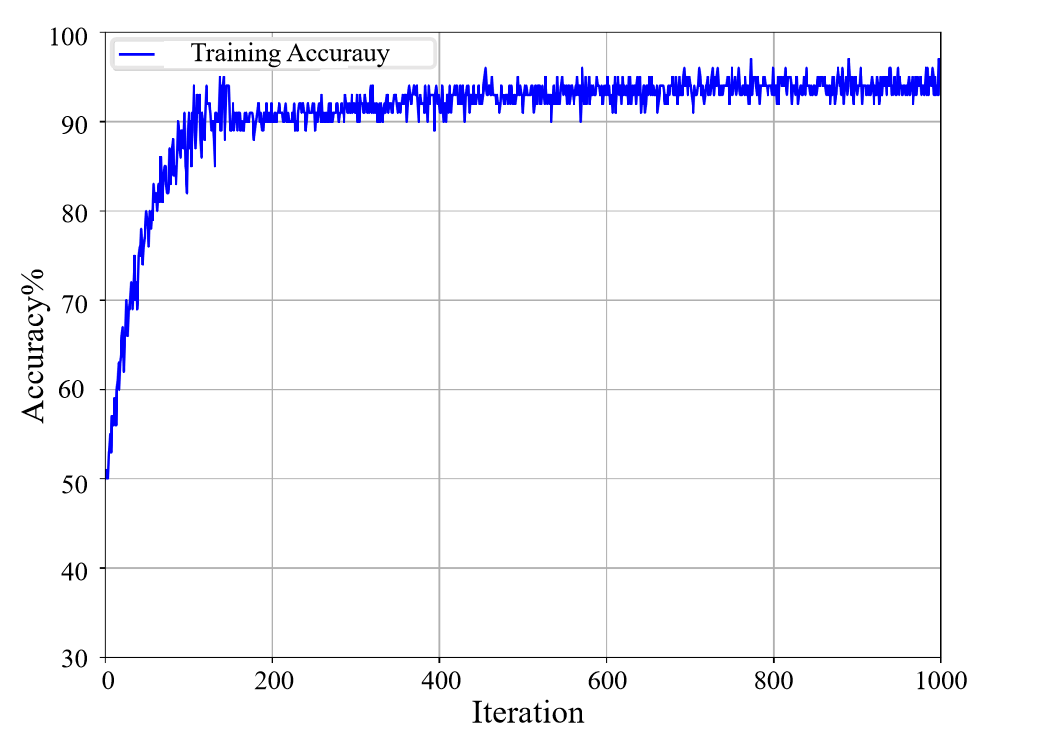}
    \caption{Training accuracy over iterations for different methods. Our method achieves nearly 100\% accuracy significantly faster than the baselines, indicating its superior learning efficiency and effectiveness.}
    \label{fig:accuracy}
\end{figure}

Tracking training accuracy is critical to assess how efficiently a model learns fault representations during training. A model that achieves high accuracy faster demonstrates its effectiveness in adapting to fault patterns. This experiment visualizes the learning process and highlights differences in training efficiency among methods. As shown in Fig.~\ref{fig:accuracy}, MMT-FD achieves nearly 100\% training accuracy significantly faster than the baseline methods, emphasizing its superior learning efficiency.

\subsection{Test on Rotor System Experimental Platform }
To further validate the effectiveness of fault diagnosis methods under complex operating conditions, we conduct experiment on the Experimental platform (Fig. \ref{fig:platform}) to conduct tests on an actual gas turbine rotor simulation platform to collect real, representative vibration data. Compared to traditional simulation or laboratory environments, the actual test platform can more accurately replicate the working conditions and potential fault states of the gas turbine, capturing complex dynamic changes and high-frequency vibration characteristics, thus providing more representative data for fault diagnosis algorithms. This not only helps verify the accuracy of theoretical methods but also ensures that the developed algorithms possess better generalization capabilities, enabling them to handle various real-world fault scenarios and operational conditions.

Specifically, firstly, we conduct a comprehensive inspection and preparation of the gas turbine rotor system platform, ensuring that the core equipment (such as the HSP2-30 high-performance servo motor) and all sensors are functioning properly. Next, we set up three typical fault states of the rotor, including the normal state, imbalance on balance disk 1, and imbalance on balance disk 2, simulating these faults by adjusting the rotor structure. Then, we install accelerometers on the exterior of the rotor and collect vibration signals at a sampling frequency of 6000 Hz to capture high-frequency vibration characteristics. Subsequently, we process the signals using an overlapping sampling method, with each sample containing 2048 data points and a sampling step size of 850 data points, ensuring the diversity and representativeness of the data. After that, we divide the collected data into training and testing sets with a 7:3 ratio, ensuring a balanced distribution of samples for each fault state in both sets. 

Following this, we monitor the experimental process in real-time to ensure the stability of the equipment and sensors, recording all relevant experimental parameters. Finally, we analyze the collected vibration data, validate and compare the performance of different fault diagnosis algorithms, assess their accuracy, robustness, and generalization capabilities, and optimize the algorithms based on the experimental results to enhance their diagnostic capabilities under real-world complex operating conditions.

\begin{figure}
    \centering
    \includegraphics[width=0.6\linewidth]{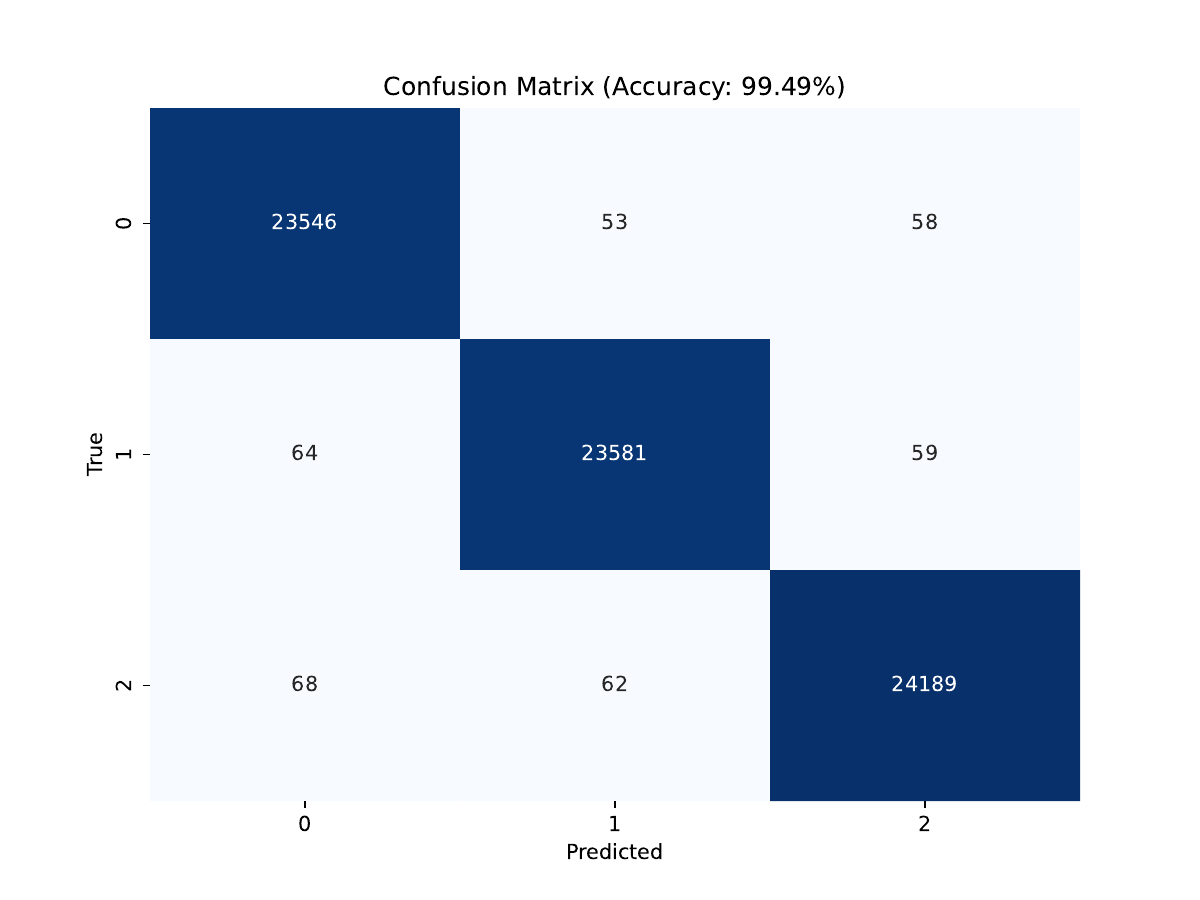}
    \caption{Fault diagnosis confusion matrix on rotor system experimental platform.}
    \label{fig:simulation_matrix}
\end{figure}

\begin{figure}
    \centering
    \includegraphics[width=0.6\linewidth]{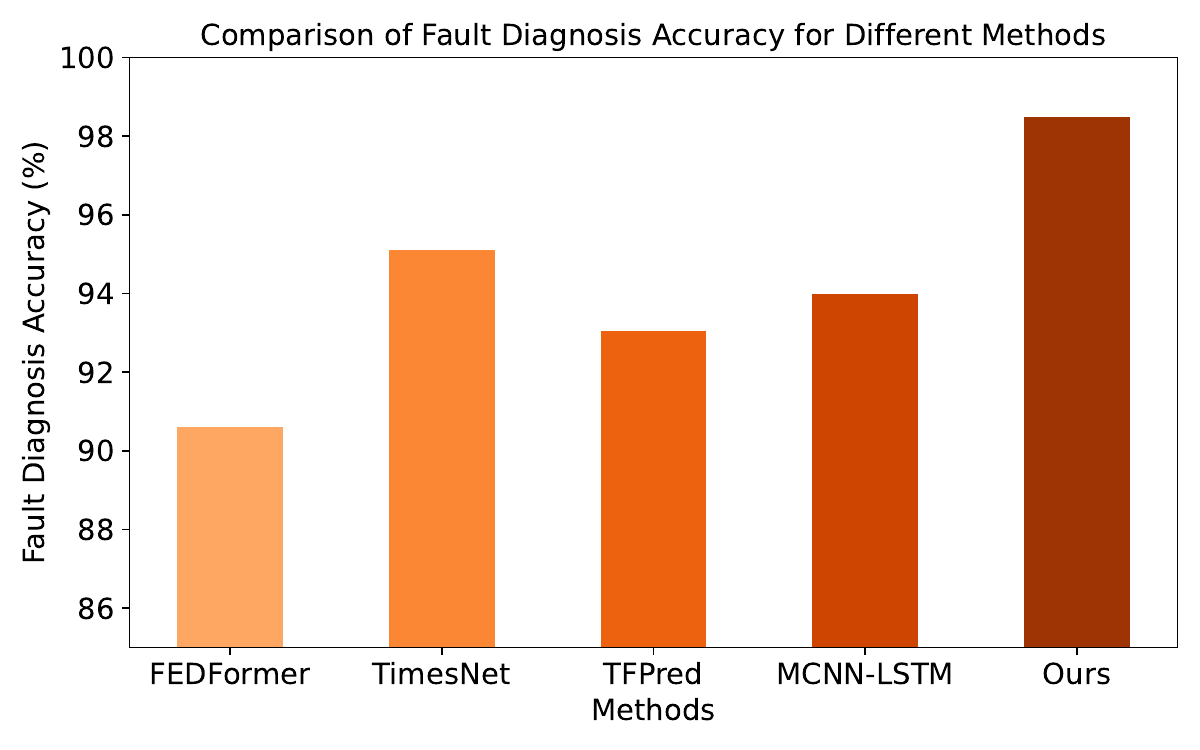}
    \caption{Comparison of fault diagnosis accuracy for different methods on rotor system experimental platform.}
    \label{fig:simulation_results}
\end{figure}

The performance and comparison results are shown in Fig. \ref{fig:simulation_matrix} and Fig. \ref{fig:simulation_results}. 
The results show that our method performs exceptionally well in the experiments, surpassing all baseline methods. Compared to the SOTA baseline, our method achieves an improvement of over 5.2\% in fault recognition accuracy. Combined with the above experimental results, our approach demonstrates significant improvements in fault diagnosis accuracy, robustness, and generalization capability, particularly in complex operating conditions, where it better adapts to different types of fault states.

\section{Conclusions}

This paper proposes MMT-FD, a framework that addresses the critical challenges of limited labeled data and lack of generalizability in fault diagnosis for rotating mechanical equipment. 
By leveraging a time-frequency domain encoder and meta-learning-based generalization, MMT-FD effectively extracts potential fault representations from unlabeled data and adapts these representations for diagnosing diverse types of mechanical equipment. The framework’s ability to perform self-supervised learning in the time-frequency domain, combined with efficient fine-tuning using minimal labeled data, highlights its practical applicability in industrial settings.
Experimental results on bearing fault and rotor test bench datasets validate the model's performance, achieving an impressive 99\% accuracy with only 1\% labeled samples. This demonstrates not only its high diagnostic accuracy but also its robust generalization capability across different mechanical systems. 
The MMT-FD framework represents a significant step forward in intelligent fault diagnosis, offering a scalable and efficient solution for real-world engineering applications where labeled data is scarce and system variability is high.

\section*{Acknowledgements}
This research is supported by Beijing Natural Science Foundation (No.IS24076) ,National Natural Science Foundation of China (No.51975058) and Beijing Natural Science Foundation(No.21JC0016).


\bibliographystyle{unsrt} 
\bibliography{example}

\end{document}